\documentclass[12pt,twoside]{article}
\makeatletter\if@twocolumn\PassOptionsToPackage{switch}{lineno}\else\fi\makeatother

      \makeatletter
\usepackage{wrapfig}

\newcounter{aubio}

\long\def\bioItem{%
\@ifnextchar[{\@bioItem}{\@@bioItem}}

\long\def\@bioItem[#1]#2#3
{
 \stepcounter{aubio}
 \expandafter\gdef\csname authorImage\theaubio\endcsname{#1}
 \expandafter\gdef\csname authorName\theaubio\endcsname{#2}
 \expandafter\gdef\csname authorDetails\theaubio\endcsname{#3}
}

\long\def\@@bioItem#1#2{
 \stepcounter{aubio}
 \expandafter\gdef\csname authorName\theaubio\endcsname{#1}
 \expandafter\gdef\csname authorDetails\theaubio\endcsname{#2}
}

\newcommand{\checkheight}[1]{%
  \par \penalty-100\begingroup%
  \setbox8=\hbox{#1}%
  \setlength{\dimen@}{\ht8}%
  \dimen@ii\pagegoal \advance\dimen@ii-\pagetotal
  \ifdim \dimen@>\dimen@ii
    \break
  \fi\endgroup}

\def\printBio{%
  \@tempcnta=0
   \loop
     \advance \@tempcnta by 1
     \def\aubioCnt{\the\@tempcnta}
     \setlength{\intextsep}{0pt}%
     \setlength{\columnsep}{10pt}%
     \newbox\boxa%
     \setbox\boxa\vbox{\csname authorDetails\aubioCnt\endcsname}
     \expandafter\ifx\csname authorImage\aubioCnt\endcsname\relax%
      \else%
       \checkheight{\includegraphics[height=1.25in,width=1in,keepaspectratio]{\csname authorImage\aubioCnt\endcsname}}
        \begin{wrapfigure}{l}{25mm}
         \includegraphics[height=1.25in,width=1in,keepaspectratio]{\csname authorImage\aubioCnt\endcsname}
        \end{wrapfigure}\par
      \fi
     {\parindent0pt\textbf{\csname authorName\aubioCnt\endcsname}\csname authorDetails\aubioCnt\endcsname \par\bigskip%
     \expandafter\ifx\csname authorImage\aubioCnt\endcsname\relax\else%
      \ifdim\the\ht\boxa < 90pt\vskip\dimexpr(90pt -\the\ht\boxa-1pc)\fi%
     \fi}
      \ifnum\@tempcnta < \theaubio
   \repeat
   }

\makeatother

\usepackage{amsfonts,amssymb,amsbsy,latexsym,amsmath,tabulary,graphicx,times,xcolor,gensymb}

\usepackage{amsmath,tabulary,graphicx,times,fancyhdr}
\usepackage{bm}
\usepackage[nonindentfirst]{titlesec}
\usepackage[utf8]{inputenc}
\usepackage[colorlinks,linkcolor=violet,citecolor=violet]{hyperref}
\usepackage{url}

\titlespacing{\section}{0pt}{.5\baselineskip}{.3\baselineskip}  
\titlespacing{\subsection}{0pt}{.5\baselineskip}{.3\baselineskip}  
\titlespacing{\subsubsection}{0pt}{.5\baselineskip}{.3\baselineskip}  
\titlespacing{\paragraph}{0pt}{.5\baselineskip}{10pt}  
\titlespacing{\subparagraph}{0pt}{.5\baselineskip}{10pt}  
\usepackage[T1]{fontenc}
\usepackage[paperheight=297mm,paperwidth=210mm,columnsep=1.93pc,right=17.5mm,left=20mm,top=20mm,bottom=20mm]{geometry}
 \date{}
\renewenvironment{abstract}{\vspace*{-1pc}\trivlist\item[]\leftskip\oupIndent\par\vskip4pt\noindent{\bfseries\abstractname}\mbox{\null}:}{\par\noindent\endtrivlist}

\makeatletter\def\oupIndent{1pt}
\def\author#1{\gdef\@author{\hskip-\dimexpr(\tabcolsep)\hskip\oupIndent\parbox{\dimexpr\textwidth-\oupIndent}{#1}}}
\def\title#1{\gdef\@title{\raggedright\bfseries\ifx\@articleType\@empty\else\@articleType\\\fi#1}}
\fancypagestyle{headings}{\fancyhf{}\fancyhead[LE]{\thepage}\fancyhead[RO]{\thepage}\fancyhead[RE]{\RunningHead}\fancyhead[LO]{\journalTitle}}
\fancypagestyle{plain}{\fancyhf{}\fancyfoot[R]{\thepage}}
\let\@articleType\@empty \def\articletype#1{\gdef\@articleType{{\normalfont\itshape#1}}}
\makeatother

\usepackage{url,multirow,morefloats,floatflt,cancel,tfrupee}
\makeatletter

\AtBeginDocument{\@ifpackageloaded{textcomp}{}{\usepackage{textcomp}}}
\makeatother
\usepackage{colortbl}
\usepackage{xcolor}
\usepackage{pifont}
\usepackage[nointegrals]{wasysym}
\makeatletter

\def\mcWidth#1{\csname TY@F#1\endcsname+\tabcolsep}

\def\cAlignHack{\rightskip\@flushglue\leftskip\@flushglue\parindent\z@\parfillskip\z@skip}
\def\rAlignHack{\rightskip\z@skip\leftskip\@flushglue \parindent\z@\parfillskip\z@skip}

\@ifundefined{etal}{}{}

\usepackage{ifxetex}
\ifxetex\else\if@twocolumn\@ifpackageloaded{stfloats}{}{\usepackage{dblfloatfix}}\fi\fi

\AtBeginDocument{
\expandafter\ifx\csname eqalign\endcsname\relax
\def\eqalign#1{\null\vcenter{\def\\{\cr}\openup\jot\m@th
  \ialign{\strut$\displaystyle{##}$\hfil&$\displaystyle{{}##}$\hfil
      \crcr#1\crcr}}\,}
\fi
}

\AtBeginDocument{%
  \@ifpackageloaded{endfloat}%
   {\renewcommand\efloat@iwrite[1]{\immediate\expandafter\protected@write\csname efloat@post#1\endcsname{}}}{\newif\ifefloat@tables}%
}%

\def\BreakURLText#1{\@tfor\brk@tempa:=#1\do{\brk@tempa\hskip0pt}}
\let\lt=<
\let\gt=>
\def\processVert{\ifmmode|\else\textbar\fi}

\@ifundefined{subparagraph}{
\def\subparagraph{\@startsection{paragraph}{5}{2\parindent}{0ex plus 0.1ex minus 0.1ex}%
{0ex}{\normalfont\small\itshape}}%
}{}

\newcommand\role[1]{\unskip}
\newcommand\aucollab[1]{\unskip}
  
\@ifundefined{tsGraphicsScaleX}{\gdef\tsGraphicsScaleX{1}}{}
\@ifundefined{tsGraphicsScaleY}{\gdef\tsGraphicsScaleY{.9}}{}
\def\checkGraphicsWidth{\ifdim\Gin@nat@width>\linewidth
	\tsGraphicsScaleX\linewidth\else\Gin@nat@width\fi}

\def\checkGraphicsHeight{\ifdim\Gin@nat@height>.9\textheight
	\tsGraphicsScaleY\textheight\else\Gin@nat@height\fi}

\def\fixFloatSize#1{}
\let\ts@includegraphics\includegraphics

\def\inlinegraphic[#1]#2{{\edef\@tempa{#1}\edef\baseline@shift{\ifx\@tempa\@empty0\else#1\fi}\edef\tempZ{\the\numexpr(\numexpr(\baseline@shift*\f@size/100))}\protect\raisebox{\tempZ pt}{\ts@includegraphics{#2}}}}

\AtBeginDocument{\def\includegraphics{\@ifnextchar[{\ts@includegraphics}{\ts@includegraphics[width=\checkGraphicsWidth,height=\checkGraphicsHeight,keepaspectratio]}}}

\DeclareMathAlphabet{\mathpzc}{OT1}{pzc}{m}{it}

\def\URL#1#2{\@ifundefined{href}{#2}{\href{#1}{#2}}}

\def\UrlOrds{\do\*\do\-\do\~\do\'\do\"\do\-}%
\g@addto@macro{\UrlBreaks}{\UrlOrds}

\edef\fntEncoding{\f@encoding}

\makeatother

\newif\ifmultipleabstract\multipleabstractfalse%
\usepackage[numbers,sort&compress,round]{natbib}

\usepackage{float}

\usepackage{listings,xcolor}

\lstdefinestyle{listing_style}{frame=single,basicstyle=\fontfamily{pcr}\selectfont,numberstyle=\tiny,xleftmargin=1pc,linewidth=.98\linewidth,backgroundcolor=\color{black!0},breaklines=true,keywordstyle=\color{blue},commentstyle=\color{darkgray},numbers=left,tabsize=3,captionpos=b,escapeinside={[@}{@]}}

\begin{document}

\def\authorCount{3}
\def\affCount{2}

\def\journalTitle{Quantitative Imaging in Medicine and Surgery}

\title{Towards Toxic and Narcotic Medication Detection with Rotated Object Detector
\thanks{The source code would be opened on https://github.com/woodywff/drug\_det\_ro}
}
\author{\textbf{Jiao Peng,\textsuperscript{{1}{$\dagger$}} Feifan Wang,\textsuperscript{{2}{$\dagger$}}  Zhongqiang Fu,\textsuperscript{{2}} Yiying Hu,\textsuperscript{{2}} Zichen Chen,\textsuperscript{{2}} Xinghan Zhou,\textsuperscript{{2}}and Lijun Wang\textsuperscript{{1}}}~\\[4pt] ~\\{$^1$Department of Pharmacy, Peking University Shenzhen Hospital, Shenzhen, China.}~\\{$^2$NuboMed, China.}~\\{$^\dagger$Co-first authors}}\def\RunningHead{Towards Toxic and Narcotic Medication Detection with Rotated Object Detector}

\maketitle 

~\\\textbf{Correspondence: }Feifan Wang~\\Email: woodywff@aliyun.com or feifan.wang@nubomed.com

\begin{abstract}
Recent years have witnessed the advancement of deep learning vision technologies and applications in the medical industry. Intelligent devices for special medication management are in great need of, which requires more precise detection algorithms to identify the specifications and locations. In this work, YOLO (You only look once) based object detectors are tailored for toxic and narcotic medications detection tasks. Specifically, a more flexible annotation with rotated degree ranging from $0^\circ$ to $90^\circ$ and a mask-mapping-based non-maximum suppression method are proposed to achieve a feasible and efficient medication detector aiming at arbitrarily oriented bounding boxes. Extensive experiments demonstrate that the rotated YOLO detectors are more suitable for identifying densely arranged drugs. The best shot mean average precision of the proposed network reaches 0.811 while the inference time is less than 300ms. \def\keywordstitle{Keywords}

\smallskip\noindent\textbf{Keywords: }{Toxic and narcotic medication, YOLO, rotated object detection\newline}
\end{abstract}
    
\section{Introduction}

Medicinal toxic drugs, narcotic drugs, psychotropic substances, and radioactive pharmaceuticals are special medications that need strict management in hospitals, especially medicinal toxic drugs and narcotic drugs. If these medications inflow into the market illegally, the harm is unimaginable, no less than “ice” (crystal methamphetamine)\unskip~\cite{peng1, peng2}. In China, these four kinds of special medications were being managed strictly by a special person, special counter, special account books, special prescription, and special book registration\unskip~\cite{peng3, peng4}. As a daily work, these medications need to clean up every day under the supervision of another staff member. In parallel, the empty ampoules of these medications after use are required to send back to the central pharmacy to ensure that the quantity of drug registration is consistent with that of empty ampoules. In the mode of traditional medicine cabinet management, many problems would be encountered. For instance, modification of anesthesia prescription is very common in clinics, which often resulted in that the records were inconsistent with the actual presence of medications in the counter, due to the delay in changing the information. As well, it is difficult to trace incidents such as internal personnel false claim, exchange, illegal prescription, residual liquid improper management \unskip~\cite{peng5, peng6}. These problems bring great risks to the management of special medication, and there is an urgent need to implement informatization management of special medication with an intelligent medicine cabinet to reduce errors, and save labor and time cost \unskip~\cite{peng7}.

As one of the cornerstone projects in computer vision with deep learning, object detection has long been attractive for researchers. The exploration and exploitation of these research achievements keep stimulating the emergence of advanced products in the industry.

Ever since AlexNet won the ImageNet 2012 competition, deep learning has returned as a leading actor in the academic and industrial world\unskip~\cite{alexnet}. Spontaneously, numerous proposals have been brought up to take advantage of deep learning in solving the object detection problem which is a combination of classification and localization tasks. Ross Girshick designed the diagram leveraging regions with convolutional neural network (R-CNN) features\unskip~\cite{rcnn}. Endeavors of such strategy include two stages, one is to extract the deep features from input images and the other is to figure out the region of interests (ROI) in which the extracted features contribute the most to the final prediction process. One intrinsic drawback of the R-CNN series is the relatively long latency, despite the advancement in its siblings like fast-RCNN\unskip~\cite{fast_rcnn} and faster-RCNN\unskip~\cite{faster_rcnn}. In 2015, Joseph Redmon came up with you only look once (YOLO)\unskip~\cite{yolo_v1} style network in which an input image is cut into different scaled grids, and in each grid, there are multiple anchors in charge of the bounding box regression. In this way, the feature embedding and re-localization on ROIs could be proceeding in one single round, which gets around the ROI searching process and tremendously decreases the time consumption. During the past few years, the YOLO series keeps evolving as the state-of-the-art object detector that not only wins titles in academic contests but also receives reputations in industrial field\unskip~\cite{yolo_v2, yolo_v3, yolo_v4, yolo_x}.  

Most of the existing annotated datasets for object detection projects provide bounding boxes as rectangles parallel to the x and y-axis, while in real-world there are specific scenarios in which objects are usually densely arranged in arbitrary orientations. Under such circumstances arose special requirements especially from the aerospace industry in which the aerial view photo is in common use.
Compared with the prediction of horizontal bounding boxes, rotated object detection has an extra variable--rotation angle $\theta$--to be anticipated. Xue Yang et.al gives out solutions in which the $\theta$ is encapsulated by specifically designed skew-IOU loss functions\unskip~\cite{yx1, yx2, yx3, yx4}. By doing this, the difference in bounding box rotated angle would contribute to the update of network parameters during the backpropagation procedure. Qi Ming et.al spare effort to invent more efficient rotated anchor learning procedures and refined feature extractors that address the inconsistency between classification and bounding box regression and make the model more suitable for the arbitrarily oriented object detection task\unskip~\cite{mq1, mq2, mq3}. Similar works for the aerial images have also been undertaken by Jiaming Han et.al, who propose the so-called oriented detection module (ODM) and feature alignment module (FAM)\unskip~\cite{hj1, hj2}.

Despite being a special case of rotated object detection, toxic and narcotic medication identification problems still need thorough exploration. 
On one hand, popular datasets used for training horizontal and rotated object detectors include few samples sharing the same distribution as the toxic and narcotic drugs of our interests. On the other hand, the existing rotated bounding box annotating method doesn't satisfy the flexibility we look forward to. Moreover, the current non-maximum suppression (NMS) for rotated object detection is still notorious as a time-consuming post-processor.

In this paper, an arbitrarily oriented object detector aimed at instantly and precisely identifying the specifications of toxic and narcotic medication is fabricated. Throughout this context, specification means a different medication or dose. The main contributions of this work boil down to three folds: 
\begin{itemize}
	\item YOLO-based backbone and head network have been revised and tuned to accomplish the toxic and narcotic medication detection target.
	\item  A more flexible rotated bounding box annotation method is devised that shrinks the rotated angle into 0 to 90 degrees and forecasts the angle value employing classification procedure.
	\item Last but not least, to make the proposed rotated toxic and narcotic medication detector feasible in real application scenarios the rotated intersection over union (IOU) is calculated with help of 0-1 mask for each found object.
\end{itemize} 

The remainder of this paper is organized as follows. Section 2 explains the details of the proposed network. Extensive experiments and comparison results analysis go into Section 3. Finally, in section 4 we conclude this work.

\section{Methodology}

\subsection{Rotated Bounding Box}
\begin{figure}[!htbp]
	\centering
	\includegraphics[width=0.65\textwidth]{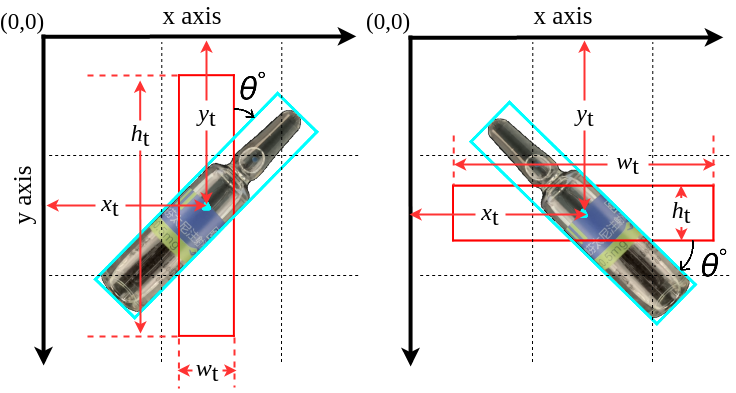}
	\caption{Illustration of rotated bounding box definition. The horizontal and rotated bounding boxes are marked in red and cyan respectively. The gray dashed lines indicate the grids in which multiple scaled and relocated anchors are given out as the predicted bounding boxes. Small value on x-y-axis goes to the up-left corner.}
	\label{theta}
\end{figure}

Previous rotated object detection researches define a bounding box with arbitrary rotated angle by common scenes that the long edge lies along the x-axis and the rotated angle ranges from -90$^{\circ}$ to 90$^{\circ}$\unskip~\cite{csl, bba_vector}. However, a rotation could have been described with degrees from 0$^{\circ}$ to 89$^{\circ}$. Instead of following the traditional long-short edge prerequisite which would impose more memory consumption, in this work, we define the rotated bounding box more flexibly. As shown in Fig.~\ref{theta}, the up-left corner is the zero point, ($x_{\rm{t}}, y_{\rm{t}}, w_{\rm{t}}, h_{\rm{t}}, \theta$) designate a rotated bounding box on the image. $x_{\rm{t}}, y_{\rm{t}}$ are coordinates of the box center point. $w_{\rm{t}}, h_{\rm{t}}$ are width and height on x-axis and y-axis respectively. ($x_{\rm{t}}, y_{\rm{t}}, w_{\rm{t}}, h_{\rm{t}}$) draws a horizontal rectangle which has nothing to do with $\theta$. The black arrow points to the positive direction of rotated angle. After the observation of 360$^\circ$ rotation it is concluded that 0$^{\circ}$ to 89$^{\circ}$ are enough to cover all the different situations as long as we change the width and height once the $\theta$ crossed the $90^{\circ}$. In Fig.~\ref{theta}, the left and right part indicate two layouts of an ampoule [jysftnzsy], both $\theta$ are $45^{\circ}$, though the two horizontal bounding boxes are in orthogonal places.     

Just like the circular smooth label\unskip~\cite{csl}, the regression of a rotated bounding box is now transformed to be the regression of a horizontal bounding box and a rotated angle $\theta$. Rather than predicting $\theta$ continuously, a classifier would be trained to decide which degree is the closest to $\theta$. Different granularity could be taken when it comes to the assignment of rotation degrees. For example, 90 granularity ends up with 1$^{\circ}$ being represented by one class label, while 180 granularity uses 180 classes to identify 90 degrees which means each class represents 0.5$^{\circ}$.

\subsection{Toxic and narcotic medication detector}
\begin{figure}[!htbp]
	\centering
	\includegraphics[width=0.8\textwidth]{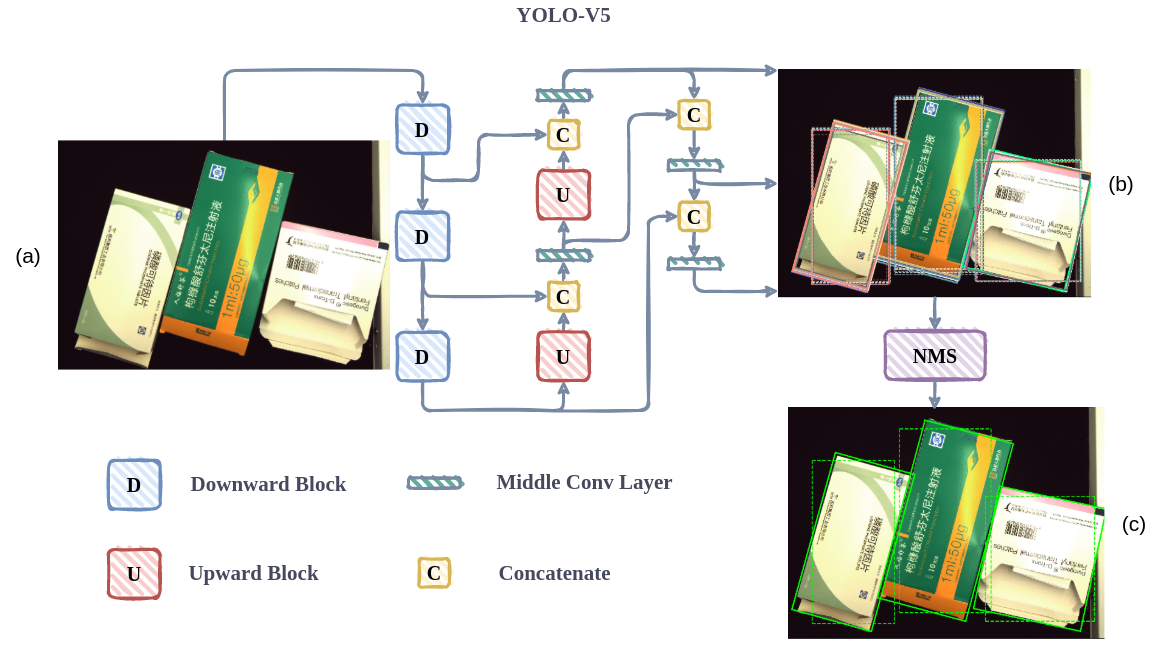}
	\caption{Schematic of the YOLO style toxic and narcotic medication detector. (a) shows the original input image. (b) is the output with all the predicted horizontal and rotated bounding boxes in dashed and solid lines respectively. The kept predictions after the NMS filter have been drawn in (c).}
	\label{model}
\end{figure}

In this work, a YOLO-based architecture is leveraged to build the rotated toxic and narcotic medication detector. As shown in Fig.~\ref{model}, an input image fed into the network would go through the downward embedding and the upward re-localization processes in turn. The essence of YOLO is that multi-scale grids are imposed on the same image and each grid gives out 3 anticipations based on the predefined anchors. For training, not all the grids participate in the forward and back-propagation procedures. Only those grids whose corresponding anchors are not too large or too small compared with the target bounding box are kept. Moreover, to enlarge the number of positive samples which are the correctly predicted bounding boxes, two more neighbor grids of each target grid are picked up and assigned with the identical target label. The backbone of this proposed architecture follows the design of YOLO-V5 small edition\unskip~\cite{yolo_v5}. 

The final output on each grid scale is a matrix $\bm{P}\in \mathbb{R}^{N_{\rm{B}}\times N_{\rm{A}}\times N_{{\rm{G}}_h}\times N_{{\rm{G}}_w}\times (5+N_{\rm{C}} + N_{\rm{D}})}$, in which $N_{\rm{B}}$ is the number of batch size, $N_{\rm{A}}$ is the number of anchors for each grid, $N_{{\rm{G}}_h}, N_{{\rm{G}}_w}$ are number of grids on y and x axis respectively, $N_{\rm{C}}$ is the number of medication specifications (classes), $N_{\rm{D}}$ is the number of degree intervals. 
Given $H, W$ as the input image height and width, the predicted horizontal rectangle based on the $j$th anchor in the $k$th row, $l$th column grid of the $i$th batch index image could be depicted by the center point coordinates$x_{\rm{p}}, y_{\rm{p}}$, the width $w_{\rm{p}}$ and the height $h_{\rm{p}}$ of the bounding box. 
\begin{align}
x_{\rm{p}} &= l W/N_{{\rm{G}}_w} + 2\sigma(P_{ijkl0}) - 0.5 \label{x_p}\\
y_{\rm{p}} &= k H/N_{{\rm{G}}_h} + 2\sigma(P_{ijkl1}) - 0.5 \label{y_p}\\
w_{\rm{p}} &= 4\sigma(P_{ijkl2})^2W_{\rm{A}_j} \label{w_p}\\
h_{\rm{p}} &= 4\sigma(P_{ijkl3})^2H_{\rm{A}_j} \label{h_p}
\end{align}
In Eq.~(\ref{x_p}$\thicksim$\ref{h_p}), $\sigma$ refers to the sigmoid function $\sigma(x)=1/(1+\rm{exp}(-x))$, $W_{\rm{A}_j}, H_{\rm{A}_j}$ are the width and height of the $j$th anchor. Let $IoU$ denote the intersection over union\unskip~\cite{ciou}, the bounding box loss value for each grid is 
\begin{align}
L\_box_{ijkl} &=
\begin{cases}
1-IoU\big((x_{\rm{p}}, y_{\rm{p}}, w_{\rm{p}}, h_{\rm{p}}), (x_{\rm{t}}, y_{\rm{t}}, w_{\rm{t}}, h_{\rm{t}})\big) & \text{if $x_{\rm{t}}$ existed,}\\
\qquad\ null & \text{otherwise.}
\end{cases}\label{l_box}
\end{align}
$L\_box$ is the mean value of all the $L\_box_{ijkl}$ who is not $null$.

Define $\bm{T}\in \mathbb{R}^{N_{\rm{B}}\times N_{\rm{A}}\times N_{{\rm{G}}_h}\times N_{{\rm{G}}_w}}$ which indicates whether or not there is an object in current grid.
\begin{align}
T_{ijkl} &=
\begin{cases}
\qquad\ 0 & \text{if $L\_box_{ijkl}$ is $null$,}\\
1-L\_box_{ijkl} & \text{otherwise.}
\end{cases}\label{}
\end{align}
The objective loss value is the binary cross entropy between $\bm{P}$ and $\bm{T}$.
\begin{equation}
L\_obj = -\frac{\sum_{ijkl}^{}{T_{ijkl}{\rm{log}}\big(\sigma(P_{ijkl4})\big) + (1-T_{ijkl}){\rm{log}}\big(1-\sigma(P_{ijkl4})\big)}}{N_{\rm{B}}N_{\rm{A}}N_{{\rm{G}}_h}N_{{\rm{G}}_w}}.
\label{}
\end{equation} 

For each target label, there is a $c_{\rm{t}}$ indicating the index of medication specification(class). Suppose there are $n_{\rm{t}}$ target boxes, let $\hat{\bm{T}}\in\mathbb{R}^{{n_{\rm{t}}}\times N_{\rm{C}}}$ be the one-hot label matrix.
\begin{align}
\hat{T}_{ij} &=
\begin{cases}
\ 1 & \text{if $j==c_{\rm{t}}$,}\\
\ 0 & \text{otherwise,}
\end{cases}\label{}
\end{align}
in which $i\in[0, n_{\rm{t}}-1], j\in[0, N_{\rm{C}}-1]$. 
The corresponding selected output could be represented as $\hat{\bm{P}}\in\mathbb{R}^{{n_{\rm{t}}}\times N_{\rm{C}}}$ each row of which are the $5$th to $(N_{\rm{C}}-1)$th items in $\bm{P}$. The classification loss value is
\begin{equation}
L\_cls = -\frac{\sum_{ij}^{}{\hat{T}_{ij}{\rm{log}}\big(\sigma(\hat{P}_{ij})\big) + (1-\hat{T}_{ij}){\rm{log}}\big(1-\sigma(\hat{P}_{ij})\big)}}{{n_{\rm{t}}} N_{\rm{C}}}.
\label{}
\end{equation} 

Similarly, to forecast the rotated degree $\theta$, a one-hot matrix $\tilde{\bm{T}}\in\mathbb{R}^{{n_{\rm{t}}}\times N_{\rm{D}}}$ is introduced as
\begin{align}
\tilde{T}_{ij} &=
\begin{cases}
\ 1 & \text{if $j==\theta N_{\rm{D}}/90$,}\\
\ 0 & \text{otherwise,}
\end{cases}\label{}
\end{align}
in which $i\in[0, n_{\rm{t}}-1], j\in[0, N_{\rm{D}}-1]$. Let $\tilde{\bm{P}}\in\mathbb{R}^{{n_{\rm{t}}}\times N_{\rm{D}}}$ be the corresponding selected rows with the $N_{\rm{C}}$th to $(N_{\rm{D}}-1)$th columns from $\bm{P}$. The loss value on degree is proposed as 
\begin{equation}
L\_\theta = -\frac{\sum_{ij}{\tilde{T}_{ij}{\rm{log}}\big(\sigma(\tilde{P}_{ij})\big) + (1-\tilde{T}_{ij}){\rm{log}}\big(1-\sigma(\tilde{P}_{ij})\big)}}{{n_{\rm{t}}} N_{\rm{D}}}.
\label{}
\end{equation} 

The total loss value throughout the whole grid scales are a weighted summary of these four kinds of losses. Given $n_{\rm{s}}$ grid scales, let $L\_box(i), L\_obj(i), L\_cls(i), L\_\theta(i)$ represent the loss values in the $i$th space, the loss in all is
\begin{equation}
L = \gamma_{\rm{box}}\sum_{i=1}^{n_{\rm{s}}}L\_box(i) + \gamma_{\rm{obj}}\sum_{i=1}^{n_{\rm{s}}}\xi_i L\_obj(i) + \gamma_{\rm{cls}}\sum_{i=1}^{n_{\rm{s}}}L\_cls(i) + \gamma_{\rm{\theta}}\sum_{i=1}^{n_{\rm{s}}}L\_\theta(i),
\label{}
\end{equation} 
in which $\xi$ denotes the weight for object loss in different grid scale, $\gamma$ is used to balance the four loss values.
\begin{table}[!htbp]
	\centering
	\caption{$IoU$ calculation for rotated bounding boxes.}\label{algorithm}\vspace{5pt}
	\begin{tabular}{cl}
		\hline
		\multicolumn{2}{l}{$\bm{get\_mask}(x, y, w, h, \theta)$:} \\
		1 & Get coordinates of up-left and bottom-right corners $x_{\rm{min}}, y_{\rm{min}}, x_{\rm{max}}, y_{\rm{max}}$ \\
		& from $x, y, w, h$\\
		2 & Let $\bm{M}\in\mathbb{N}^{h_{\rm{m}}\times w_{\rm{m}}}$ be the mask, $M_{ij}=0, i\in[0, h_{\rm{m}}), j\in[0, w_{\rm{m}}]$\\
		3 & \textbf{if} $\theta==0$:\\
		4 & \quad $M_{ij}=1, i\in[y_{\rm{min}}, y_{\rm{max}}), 
		j\in[x_{\rm{min}}, x_{\rm{max}})$\\
		5 & \textbf{else}:\\
		6 & \quad Get slopes and intercepts of rectangle edges, 
		say $a_0, b_{00}, b_{01}, a_1, b_{10}, b_{11}$ \\
		7 & \quad $M_{ij}=1, i,j\in(i>a_0 j+b_{00}) \& (i>a_0 j + b_{01})$\\
		& \quad $\&(i>a_1 j + b_{10}) \& (i<a_1 j + b_{11}) $\\
		8 & \textbf{return} $\bm{M}$\\
		
		\multicolumn{2}{l}{$\bm{iou\_ro}(x_0, y_0, w_0, h_0, \theta_0, x_1, y_1, w_1, h_1, \theta_1)$:} \\
		1 & $\bm{M\_0} = \bm{get\_mask}(x_0, y_0, w_0, h_0, \theta_0)$,
		$\bm{M\_0}\in\mathbb{N}^{h_{\rm{m}}\times w_{\rm{m}}}$\\
		2 & $\bm{M\_1} = \bm{get\_mask}(x_1, y_1, w_1, h_1, \theta_1)$,
		$\bm{M\_1}\in\mathbb{N}^{h_{\rm{m}}\times w_{\rm{m}}}$\\
		3 & Get intersection $I = \sum_{i}^{h_{\rm{m}}}\sum_{j}^{w_{\rm{m}}}M\_0_{ij}M\_1_{ij}$.\\
		4 & Get union $U = \sum_{i}^{h_{\rm{m}}}\sum_{j}^{w_{\rm{m}}}M\_0_{ij} + \sum_{i}^{h_{\rm{m}}}\sum_{j}^{w_{\rm{m}}}M\_1_{ij}$\\
		5 & $IoU = I/U$\\
		6 & \textbf{return} $IoU$\\
		\hline
	\end{tabular}
\end{table}

\subsection{Masked NMS}
One of the most challenging tasks when designing an object detector is how to impose the NMS filter on the found boxes. There is no difference between horizontal and rotated bounding box NMS except for the $IoU$ calculation\unskip~\cite{ciou}. For horizontal $IoU$ it is not complicated to figure out the rectangle areas, while for rotated $IoU$ the intersection and union areas are irregular polygons which are not easy to get. Pure Cartesian geometry is a clear way to get the answer but it requires extremely high professional expertise. 

Alternatively, in this work, we devise an approximation leveraging 0-1 masks to find the $IoU$ between two rotated bounding boxes.
The idea is intrinsically straightforward and similar to former works like the mask-scoring\unskip~\cite{mask_scoring} and the PIOU\unskip~\cite{piou}. Table~\ref{algorithm} illustrates the mechanism of our proposed solution. Given two boxes depicted by the quintet $(x, y, w, h, \theta)$, the mask could be received utilizing $get\_mask$, and the $iou\_ro$ is in charge of getting $IoU$ between these two rotated boxes. $h_{\rm{m}}, w_{\rm{m}}$ are hyperparameters deciding the mask size.

\section{Implementation and Result}
\subsection{Dataset and Preprocessing}
There are 38 toxic and narcotic medication specifications in Peking University Shenzhen Hospital. To alleviate the pressure from manual annotation, we randomly mix the single specification images to generate more annotated samples. The random manners include rotation, resize, perspective transformation, random background, random location, and repetition. Because the source images used for data generation are caught in unrestricted circumstances which may be distinct to the environment of our product, we separate the training process into two stages. In the first stage, 6000 generated images together with 523 manually annotated photos are fed into the model to get pre-trained weights. Then the pre-trained model is fine-tuned on the 523 images again as the second stage. 42 manually annotated photos are left for inference experiments. Each object has both horizontal and rotated bounding box labels.   

All the images in the dataset are in RGB format with 0 to 255-pixel values. The shape of input data is 640$\times$640. Each sample would go through the 255 division and the min-max normalization individually. When it comes to data augmentation, we exert a series of configurable maneuvers including mosaic transformation, moving, shrinking, rotation, cropping, horizontal and vertical flipping, augmentation on hue, saturation, and value. For bounding box augmentation we do not leverage the rotation process while for rotated bounding box the shrink ratios on the x and y-axis are the same.

\subsection{Configuration} 
This work has been developed under the environment of a single RTX 3060Ti GPU card and PyTorch framework. As mentioned above, the backbone architecture is following the YOLO-V5-S\unskip~\cite{yolo_v5}. Number of grid scales $n_{\rm{s}}=3$, the respective grid steps are 8, 16, 32 on both axis. Correspondingly, the objective loss weight $\xi$s are 4, 1, 0.4. The balance parameters $\gamma_{\rm{box}}=0.05, \gamma_{\rm{cls}}=0.5, \gamma_{\rm{obj}}=1, \gamma_{\rm{\theta}}=0.5$. Target box edges longer than 4 times or smaller than 1/4 times of anchor edge would be ignored. $180^\circ$ and $90^\circ$ angel granularity have been tested. For comparison, we have trained both models in horizontal and rotated versions. There are 200 epochs training on the 6000+523 dataset and another 200 epochs on the 523 datasets. The batch size is 1 for training and validation. Optimizer is Adam\unskip~\cite{adam} starting with learning rate of 1e-3. The optimizer scheduler decreases the learning rate to 50\% once no improvement was seen in the loss value of 10 epochs. For NMS, the confidence threshold is 0.45 for both horizontal and rotated scenarios, the IoU thresholds are 0.45 and 0.25 respectively.

\subsection{Result Analysis}\label{section}
In this section, comparisons have been undertaken on Yolo-v5 tailored for toxic and narcotic medication detection tasks and the proposed arbitrary oriented bounding box detectors with $\theta=90$ and $\theta=180$ respectively. Firstly, the influence of mask size $h_{\rm{m}}, w_{\rm{m}}$ on the performance of models are unveiled.

\begin{figure}[!htbp]
	\centering
	\includegraphics[width=0.55\textwidth]{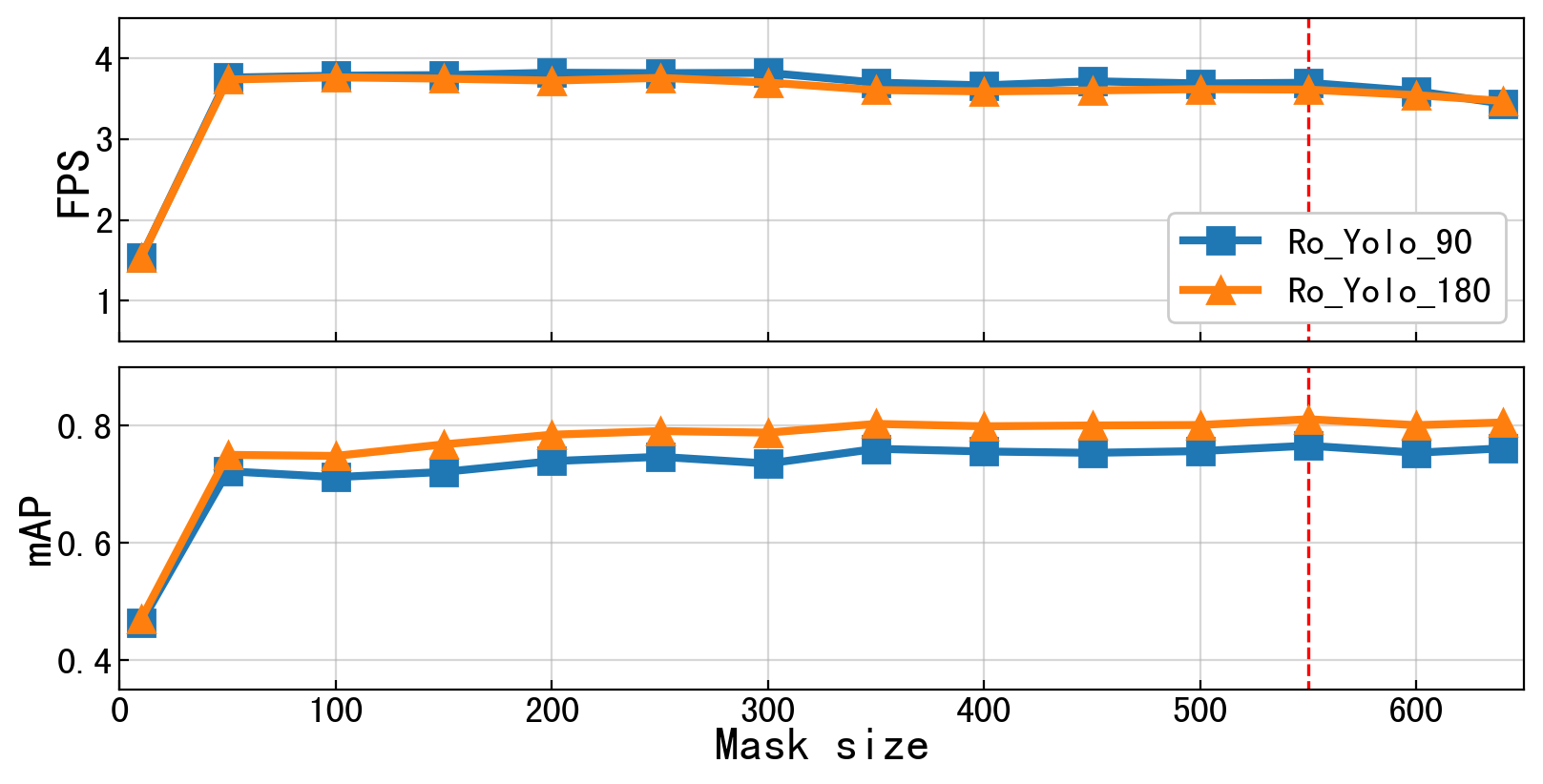}
	\caption{Comparison results with different mask sizes. The two rotated degree modes are indicated as `Ro\_Yolo\_90' and `Ro\_Yolo\_180'. The red dashed line indicates the chosen mask size where $h_{\rm{m}}=w_{\rm{m}}=550$.}
	\label{fps_map_masks}
\end{figure}

Fig.~\ref{fps_map_masks} exhibits the evaluation with the inference dataset on multiple mask sizes ($h_{\rm{m}}=w_{\rm{m}}$). Both inference speed in terms of frame per second (FPS) and accuracy in terms of mean average precision (mAP)\footnote{\scriptsize{Throughout this work, we follow the COCO dataset\unskip~\cite{coco} tradition to calculate average precisions on different IoU thresholds ranging from 0.5 to 0.95 with 0.05 step. $\rm{AP}_{50}, \rm{AP}_{75}, \rm{AP}_{95}$ refer to the mean value of average precision for the whole classes when IoU is 0.5, 0.75, and 0.95 in respective. mAP is an average of all APs.}} show a jump at the beginning and stays calm for the rest. This phenomenon is caused by the fact that small masks, despite coming with less memory reservation, would result in more ambiguous overlapped bounding boxes. Consequently, we choose $h_{\rm{m}}=w_{\rm{m}}=550$ as default for the testing on two rotated Yolo models with different $\theta$s.

\begin{figure}[!htbp]
	\centering
	\includegraphics[width=0.7\textwidth]{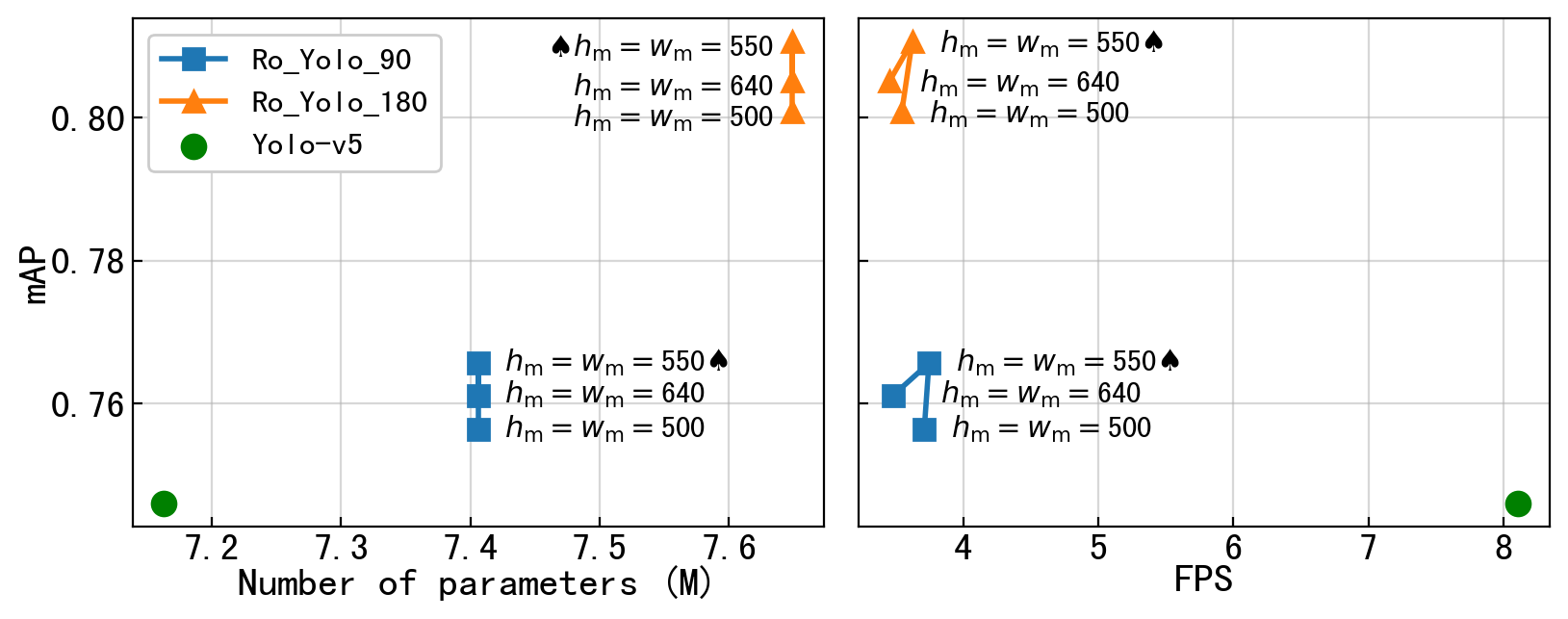}
	\caption{Performance comparison on different models. Green circles represent tailored Yolo-V5, blue squares indicate rotated Yolo with $\theta=90$, orange triangles are rotated Yolo with $\theta=180$. Three mask sizes have been illustrated in this figure, and the black spade highlights the chosen configuration.}
	\label{map_nparam_fps}
\end{figure}
The tradeoff between accuracy and efficiency has been illustrated in Fig.~\ref{map_nparam_fps}, in which the model size measured by the number of parameters, FPS, and mAP on different configured models are presented. 
The pros and cons of horizontal and arbitrary oriented object detectors are obvious to figure out. The best shot mAP of rotated Yolo model with $\theta=180$ and $h_{\rm{m}}=w_{\rm{m}}=550$ outperforms that of Yolo-V5 more than 5 percents. On the other hand, rotated detectors have hundreds of thousands of more parameters than their counterparts and need a longer time for inference. 

\begin{figure}[!htbp]
	\centering
	\includegraphics[width=0.7\textwidth]{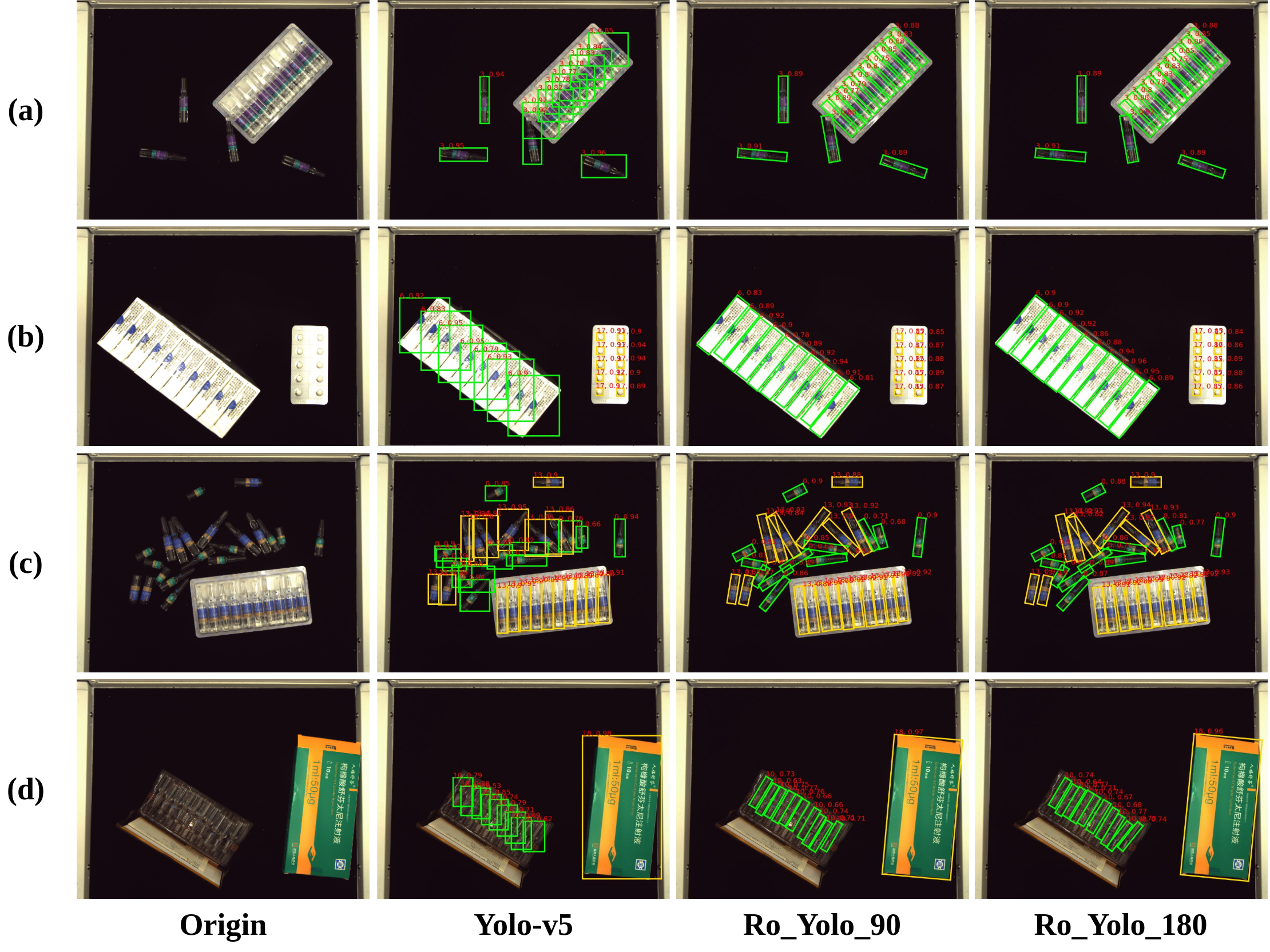}
	\caption{Detected results of different toxic and narcotic meditation detectors. The horizontal and rotated rectangles indicate the identified (brand-new and used) ampoules, pills, and boxes. In each photo, the same color represents one specification. The red labels mark the specification classes and the confidence ratio.}
	\label{images}
\end{figure}

In Fig.~\ref{images} we give out a few examples of the toxic and narcotic medications accompanied by the detected results. Although both the Yolo-v5 and rotated Yolo detectors could figure out most of the medication specifications correctly, it is easy to see the rotated Yolo detectors are better at identifying densely arranged drugs. In conditions of (a), (b), and (d), when the drugs in parallel are placed in a larger inclination angle, Yolo-v5 is hardly to precisely mark all the targets partly because there are large intersections existing among horizontal rectangles. Since most ampoules encapsulated injections are put in plastic trays, for instance, the sufentanil citrate injection (blue ampoule) and the pethidine hydrochloride injection (purple ampoule) as shown in Fig.~\ref{images}, the rotated Yolo detectors could provide more feasible predictions.

Table.~\ref{table} exhibits more details about the contrast test, in which the rotated models are configured with $h_{\rm{m}}=w_{\rm{m}}=550$. 
\begin{table}[!htbp]
	\centering
	\caption{Contrast test results on different models.}\label{table}\vspace{5pt}
	\begin{tabular}{cccccccccc}
		\hline
		\multicolumn{3}{c}{Model} & $\rm{mAP}$ & $\rm{AP_{50}}$ & $\rm{AP_{75}}$ & $\rm{AP_{95}}$ & Precision & Recall & F1 \\
		\hline
		\multicolumn{3}{c}{Yolo-v5} & 0.728 & 0.924 & 0.798 & 0.227 & 0.801 & 0.648 & 0.7\\ 
		\multicolumn{3}{c}{Ro\_Yolo\_90} & 0.766 & 0.97 & 0.817 & 0.222 & 0.796 & 0.656 & 0.693\\
		\multicolumn{3}{c}{Ro\_Yolo\_180} & 0.811 & 0.984 & 0.926 & 0.267 & 0.813 & 0.689 & 0.722\\
		\hline
	\end{tabular}
\end{table}
As the footnote in Section.~\ref{section} explained, the average precision when IoU is 0.5, 0.75, and 0.95 are listed. mAP is the mean value all over the APs. Moreover, we also calculate the precision, recall, and F1 values. Without be distinguished as different objects, the whole classes of medications are taken into account to get a single value for each confidence and IoU threshold. Further, the mean value throughout these single values is kept as what has been presented in Table.~\ref{table}. The mAPs are in accordance with that of Fig.~\ref{map_nparam_fps} which demonstrates the advantage of rotated detectors in terms of accuracy. When the IoU threshold increased, the average precision of rotated models, especially the one with $\theta=180$, deteriorates slower than Yolo-v5. According to precision, recall, and F1 values we could find that rotated Yolo detectors outperform Yolo-v5 mostly in recall which indicates the rotated models could find more bounding boxes correctly. The same evidence has been proven in Fig.~\ref{images}.

\section{Conclusions}
In this paper, the YOLO-based object detectors for toxic and narcotic meditation detection have been developed. With the more flexible rotated bounding box annotation method and the mask NMS at hand, we propose a medication detection network with arbitrarily oriented bounding boxes. Extensive contrast tests have demonstrated the feasibility and efficiency of the designed models.

Recently, new trends of deep learning advancement in object detection have attracted academic and industrial attention. For example, the successfully cross-field proved transformer has been transformed to fit the object detection task\unskip~\cite{swin_transformer}. Hinton's group demonstrates the feasibility of solving the object detection as an image captioning problem\unskip~\cite{pix2seq}. 
All these inspiring innovations shed light on the possibility for us to provide more effective and efficient products for the medical application community.

\section*{Acknowledgments}
This work was kindly funded by the National Natural Science Foundation of China (81660479), Bethune Charitable Foundation (B-19-H-20200622), Medical Science and Technology Research Foundation of Guangdong province, China (A2020157), and Peking University Shenzhen Hospital (JCYJ2018009).   

\bibliographystyle{vancouver}

\bibliography{article}

\begin{thebibliography}{10}

\bibitem{peng1}
Hall W.
\newblock The future of the international drug control system and national drug
  prohibitions.
\newblock Addiction. 2018;113(7):1212--1223.
\newblock Available from: \url{https://doi.org/10.1111/add.13941}.

\bibitem{peng2}
Rozenek EB, Wilczy\'{n}ska K, Monika G, Napoleon W.
\newblock Designer drugs still a threat?
\newblock Przegl Epidemiol. 2019;73(3):337--347.
\newblock Available from: \url{https://doi.org/10.32394/pe.73.23}.

\bibitem{peng3}
State Council of the People's Republic of China. Regulations on the Control of
  Narcotic Drugs and Psychotropic Substances. 2005;8-3.

\bibitem{peng4}
Chen S, Zhen C, Shi L.
\newblock Study on the changing process of the narcotic drugs and psychotropic
  substances catalogues of China (1949-2019). 2021;30(11):989--996.

\bibitem{peng5}
Shuai L, Luo X, Xinggang MA.
\newblock Application Effect and Experience of Hospital Narcotic Drug
  Information Management System.
\newblock Chinese Journal of Health Informatics and Management.
  2019;16(5):618--620.

\bibitem{peng6}
Wang X, Yan D, Hospital JC.
\newblock Application of Multifunctional Narcotic Drug Management Cabinets in
  Operating Rooms.
\newblock Pharmaceutical and Clinical Research. 2018;26(4):318--320.

\bibitem{peng7}
Zheng W, Pan O.
\newblock Application and Practice of Poison and Hemp Based on Integrated
  Platform and Internet of Things.
\newblock China Digital Medicine. 2017;12(4):97--99.

\bibitem{alexnet}
Krizhevsky A, Sutskever I, Hinton GE.
\newblock Imagenet classification with deep convolutional neural networks.
  2012;p. 84--90.
\newblock Available from: \url{https://doi.org/10.1145/3065386}.

\bibitem{rcnn}
Girshick R, Donahue J, Darrell T, Malik J.
\newblock Rich Feature Hierarchies for Accurate Object Detection and Semantic
  Segmentation.
\newblock In: Proceedings of the IEEE/CVF Conference on Computer Vision and
  Pattern Recognition (CVPR); 2014. p. 580--587.
\newblock Available from: \url{https://doi.org/10.1109/CVPR.2014.81}.

\bibitem{fast_rcnn}
Girshick R.
\newblock Fast R-CNN.
\newblock In: 2015 IEEE International Conference on Computer Vision (ICCV);
  2015. p. 1440--1448.
\newblock Available from: \url{https://doi.org/10.1109/ICCV.2015.169}.

\bibitem{faster_rcnn}
Ren S, He K, Girshick R, Sun J.
\newblock Faster R-CNN: Towards Real-Time Object Detection with Region Proposal
  Networks.
\newblock IEEE Transactions on Pattern Analysis and Machine Intelligence.
  2017;39(6):1137--1149.
\newblock Available from: \url{https://doi.org/10.1109/TPAMI.2016.2577031}.

\bibitem{yolo_v1}
Redmon J, Divvala S, Girshick R, Farhadi A.
\newblock You Only Look Once: Unified, Real-Time Object Detection.
\newblock In: Proceedings of the IEEE/CVF Conference on Computer Vision and
  Pattern Recognition (CVPR); 2016. p. 779--788.
\newblock Available from: \url{https://doi.org/10.1109/CVPR.2016.91}.

\bibitem{yolo_v2}
Redmon J, Farhadi A.
\newblock YOLO9000: Better, Faster, Stronger.
\newblock In: Proceedings of the IEEE/CVF Conference on Computer Vision and
  Pattern Recognition (CVPR); 2017. p. 6517--6525.
\newblock Available from: \url{https://doi.org/10.1109/CVPR.2017.690}.

\bibitem{yolo_v3}
Redmon J, Farhadi A.
\newblock Yolov3: An incremental improvement.
\newblock arXiv preprint arXiv:180402767. 2018;Available from:
  \url{https://arxiv.org/abs/1804.02767}.

\bibitem{yolo_v4}
Bochkovskiy A, Wang CY, Liao HYM.
\newblock YOLOv4: Optimal Speed and Accuracy of Object Detection.
\newblock arXiv preprint arXiv:200410934. 2020;Available from:
  \url{https://arxiv.org/abs/2004.10934v1}.

\bibitem{yolo_x}
Ge Z, Liu S, Wang F, Li Z, Sun J.
\newblock YOLOX: Exceeding YOLO Series in 2021.
\newblock arXiv preprint arXiv:210708430. 2021;Available from:
  \url{https://arxiv.org/abs/2107.08430}.

\bibitem{yx1}
Yang X, Yang J, Yan J, Zhang Y, Zhang T, Guo Z, et~al.
\newblock SCRDet: Towards More Robust Detection for Small, Cluttered and
  Rotated Objects.
\newblock In: IEEE/CVF International Conference on Computer Vision (ICCV);
  2019. p. 8231--8240.
\newblock Available from: \url{https://doi.org/10.1109/ICCV.2019.00832}.

\bibitem{yx2}
Yang X, Liu Q, Yan J, Li A, Zhang Z, Yu G.
\newblock R3Det: Refined Single-Stage Detector with Feature Refinement for
  Rotating Object.
\newblock arXiv preprint arXiv:190805612. 2019;Available from:
  \url{https://arxiv.org/abs/1908.05612}.

\bibitem{yx3}
Qian W, Yang X, Peng S, Yan J, Guo Y.
\newblock Learning Modulated Loss for Rotated Object Detection.
\newblock Proceedings of the AAAI Conference on Artificial Intelligence. 2021
  May;35(3):2458--2466.
\newblock Available from:
  \url{https://ojs.aaai.org/index.php/AAAI/article/view/16347}.

\bibitem{yx4}
Yang X, Yang X, Yang J, Ming Q, Wang W, Tian Q, et~al.
\newblock Learning High-Precision Bounding Box for Rotated Object Detection via
  Kullback-Leibler Divergence.
\newblock arXiv preprint arXiv:210601883. 2021;Available from:
  \url{https://arxiv.org/abs/2106.01883}.

\bibitem{mq1}
Ming Q, Zhou Z, Miao L, Zhang H, Li L.
\newblock Dynamic Anchor Learning for Arbitrary-Oriented Object Detection.
\newblock Proceedings of the AAAI Conference on Artificial Intelligence. 2021
  May;35(3):2355--2363.
\newblock Available from:
  \url{https://ojs.aaai.org/index.php/AAAI/article/view/16336}.

\bibitem{mq2}
Ming Q, Miao L, Zhou Z, Dong Y.
\newblock CFC-Net: A Critical Feature Capturing Network for Arbitrary-Oriented
  Object Detection in Remote-Sensing Images.
\newblock IEEE Transactions on Geoscience and Remote Sensing. 2021;p. 1--14.
\newblock Available from: \url{https://doi.org/10.1109/TGRS.2021.3095186}.

\bibitem{mq3}
Ming Q, Miao L, Zhou Z, Song J, Yang X.
\newblock Sparse Label Assignment for Oriented Object Detection in Aerial
  Images.
\newblock Remote Sensing. 2021;13(14).
\newblock Available from: \url{https://www.mdpi.com/2072-4292/13/14/2664}.

\bibitem{hj1}
Han J, Ding J, Li J, Xia GS.
\newblock Align Deep Features for Oriented Object Detection.
\newblock IEEE Transactions on Geoscience and Remote Sensing. 2021;p. 1--11.
\newblock Available from: \url{https://doi.org/10.1109/TGRS.2021.3062048}.

\bibitem{hj2}
Han J, Ding J, Xue N, Xia GS.
\newblock ReDet: A Rotation-Equivariant Detector for Aerial Object Detection.
\newblock In: Proceedings of the IEEE/CVF Conference on Computer Vision and
  Pattern Recognition (CVPR); 2021. p. 2786--2795.
\newblock Available from: \url{https://arxiv.org/abs/2103.07733}.

\bibitem{csl}
Yang X, Yan J.
\newblock Arbitrary-Oriented Object Detection with Circular Smooth Label.
\newblock In: Proceedings of the European Conference on Computer Vision (ECCV);
  2020. p. 677--694.
\newblock Available from: \url{https://doi.org/10.1007/978-3-030-58598-3_40}.

\bibitem{bba_vector}
Yi J, Wu P, Liu B, Huang Q, Qu H, Metaxas D.
\newblock Oriented Object Detection in Aerial Images With Box Boundary-Aware
  Vectors.
\newblock In: Proceedings of the IEEE/CVF Winter Conference on Applications of
  Computer Vision (WACV); 2021. p. 2150--2159.
\newblock Available from: \url{https://arxiv.org/abs/2008.07043v2}.

\bibitem{yolo_v5}
Glenn J, et~al.
\newblock YOLO-V5.
\newblock https://githubcom/ultralytics/yolov5. 2021;Available from:
  \url{https://github.com/ultralytics/yolov5}.

\bibitem{ciou}
Zheng Z, Wang P, Liu W, Li J, Ye R, Ren D.
\newblock Distance-IoU loss: Faster and better learning for bounding box
  regression.
\newblock In: Proceedings of the AAAI Conference on Artificial Intelligence.
  vol.~34; 2020. p. 12993--13000.
\newblock Available from: \url{https://arxiv.org/abs/1911.08287v1}.

\bibitem{mask_scoring}
Huang Z, Huang L, Gong Y, Huang C, Wang X.
\newblock Mask Scoring R-CNN.
\newblock In: Proceedings of the IEEE/CVF Conference on Computer Vision and
  Pattern Recognition (CVPR); 2019. p. 6402--6411.
\newblock Available from: \url{https://doi.org/10.1109/CVPR.2019.00657}.

\bibitem{piou}
Chen Z, Chen K, Lin W, See J, Yu H, Ke Y, et~al.
\newblock PIoU Loss: Towards Accurate Oriented Object Detection in Complex
  Environments.
\newblock In: Proceedings of the European Conference on Computer Vision (ECCV);
  2020. p. 195--211.
\newblock Available from: \url{https://doi.org/10.1007/978-3-030-58558-7_12}.

\bibitem{adam}
Kingma DP, Ba J.
\newblock Adam: A method for stochastic optimization.
\newblock arXiv preprint arXiv:14126980. 2014;Available from:
  \url{https://arxiv.org/abs/1412.6980}.

\bibitem{coco}
Lin TY, Maire M, Belongie S, Hays J, Perona P, Ramanan D, et~al.
\newblock Microsoft COCO: Common Objects in Context.
\newblock In: Computer Vision -- ECCV 2014. Cham: Springer International
  Publishing; 2014. p. 740--755.
\newblock Available from: \url{https://doi.org/10.1007/978-3-319-10602-1_48}.

\bibitem{swin_transformer}
Liu Z, Lin Y, Cao Y, Hu H, Wei Y, Zhang Z, et~al.
\newblock Swin transformer: Hierarchical vision transformer using shifted
  windows.
\newblock arXiv preprint arXiv:210314030. 2021;Available from:
  \url{https://arxiv.org/abs/2103.14030}.

\bibitem{pix2seq}
Chen T, Saxena S, Li L, Fleet DJ, Hinton G.
\newblock Pix2seq: A Language Modeling Framework for Object Detection.
\newblock arXiv preprint arXiv:210910852. 2021;Available from:
  \url{https://arxiv.org/abs/2109.10852}.

\end{thebibliography}

\section*{Author biography}

\bioItem[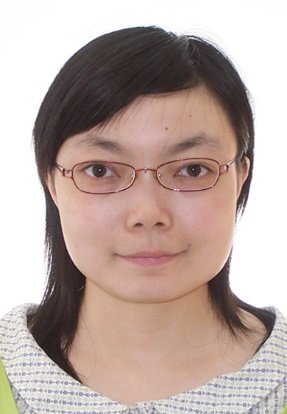]{Jiao Peng}{~is a senior pharmacist at the Peking University Shenzhen Hospital, and she graduated in Pharmacology from Li Ka Shing Faculty of Medicine, the University of Hong Kong, with a PhD degree in 2015. Her research interests include intelligent management of hospital drugs, safety monitoring and vigilance of drugs in clinics, etc.}

\bioItem[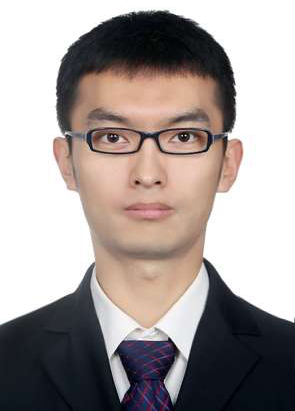]{Feifan Wang}{~received his Ph.D. degree in control science and engineering from the School of Automation, Beijing Institute of Technology in 2018. From 2018 to 2020 he had been working on brain tumor segmentation with deep learning in the Key Laboratory for NeuroInformation of Ministry of Education, School of Life Science and Technology, University of Electronic Science and Technology of China. He is currently a researcher in NuboMed. His research interests include object detection, neural architecture search, brain magnetic resonance imaging processing, etc.}

\bioItem[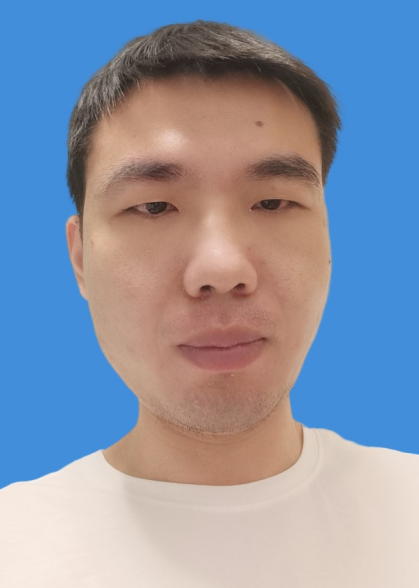]{Zhongqiang Fu}{~received his master's degree in power engineering from  Harbin Institute of Technology, Shenzhen in 2017. His current research interests include deep learning and computer vision for the Internet of Things and healthcare.}

\bioItem[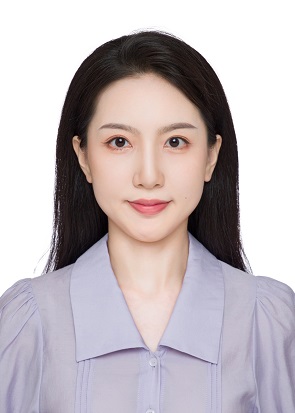]{Yiying Hu}{~received the master's degree in Information and Communication Engineering from the College of Information Engineering, East China Jiaotong University.  She currently works for NuboMed. Her research is focused on object detection and medical image processing, etc.}

\bioItem[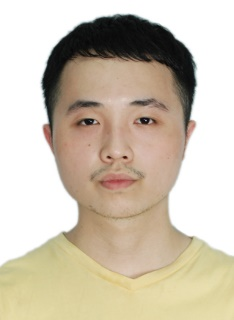]{Zichen Chen}{~received his Bio-Engineering bachelor's degree from Fujian Agriculture and Forestry University in 2014. He is now working at NuboMed as a Python Engineer in charge of data analysis and machine learning application development.}

\bioItem[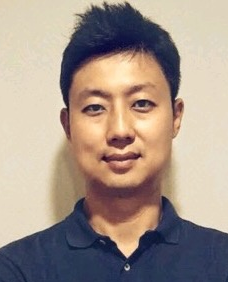]{Xinghan Zhou}{~graduated from the University of California Irvine with a Bachelor of Science in Electrical and Electronics Engineering in 2005. He is the founder and CEO of NuboMed which is a leading provider of complete solutions for the Internet of Things smart hospitals in China. Since 2012, he has initiated multiple research and development centers in Shenzhen, Hefei, and California to engage in continuously supplying artificial intelligent medical solutions for the community.}

\bioItem[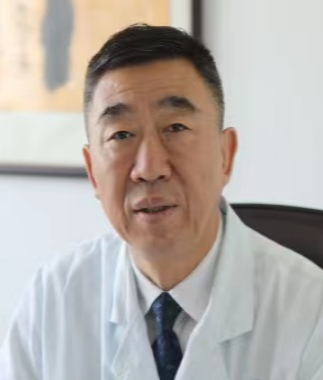]{Lijun Wang}{~is a senior pharmacist and the director of Department of Pharmacy, Peking University Shenzhen Hospital. He graduated in Chinese medicine from School of Pharmaceutical Sciences, Changchun University of Chinese Medicine, with a Bachelor degree in 1986. His research interests include pharmacy management, intelligent management of hospital drugs, safety monitoring and vigilance of drugs in clinics, etc.}

\printBio 
\end{document}